\pdfoutput=1
\documentclass[11pt]{article}
\usepackage[preprint]{acl}
\usepackage{times}
\usepackage{latexsym}
\usepackage[T1]{fontenc}
\usepackage[utf8]{inputenc}
\usepackage{microtype}
\usepackage{inconsolata}
\usepackage{graphicx}
\usepackage{booktabs}
\usepackage{amsmath,amssymb,amsthm}
\usepackage{multirow}
\usepackage{array}
\usepackage{xcolor}
\usepackage{tikz}
\usetikzlibrary{arrows.meta,positioning,fit}

\newtheorem{proposition}{Proposition}
\newcommand{\method}{\textsc{Lineup}}

\title{Contagious Fingerprints: When LLMs Inherit Another Model's Identity}

\author{
  Zailong Tian \\
  Nanyang Technological University \\
  \texttt{ztian004@e.ntu.edu.sg}
}
\begin{document}
\maketitle

\begin{abstract}
The fingerprint by which a classifier recognizes the LLM behind a text is usually treated as a fixed property of the model. We show that it is contagious: a model that continues a conversation begun by another model inherits that model's fingerprint. This undermines audits of API relays, which can serve the first turns of a conversation with the model a customer pays for and silently hand the rest to a cheaper one. After such a handoff, detection of the substitute by a text classifier drops from 0.97 to 0.72 (DeepSeek replacing Claude Sonnet) and from 0.87 to 0.54 for a same-vendor downgrade (Haiku), and retraining on handoff data repairs only substitutes seen in training. Preregistered experiments trace the contagion to imitation of the history's style. History also moves a model away from its own fingerprint: active probes are not camouflaged, but when calibrated on fresh conversations they flag \emph{honest} models on long contexts at rates up to 1.00, and a standard MMD equality test rejects all nine models we test. Both failures come from comparing a history-dependent reply with a history-free reference. We propose \method{}, which samples references from the claimed model \emph{on the same history} and tests the audited reply by its rank among them. \method{} controls false positives by exchangeability and needs no fitted threshold or log-probabilities. With contemporaneous references on two closed models, its false-positive rate is 0.056 (preregistered bound 0.10), and it detects 0.81 of substitute sessions across six pairs after four audited turns.
\end{abstract}

\section{Introduction}

LLMs leave fingerprints. Classifiers recover the source model of a text with high accuracy \citep{sun2025idiosyncrasies}, and audits of LLM APIs build on the same regularities: to check that a provider serves the model it advertises, they compare its answers with a reference collected from the genuine model \citep{cai2025substitution,gao2025met,zhu2025rut,pasquini2025llmmap,zhang2026iris}. These methods share an assumption that is rarely stated: the fingerprint belongs to the model. The reference is collected once, in fresh contexts, and is expected to describe the model in any conversation it takes part in.

We show that this assumption fails in multi-turn conversations, because fingerprints are \emph{contagious}. When a model continues a conversation whose earlier turns were written by another model, its replies move toward that model's fingerprint: the logit that a neural attribution classifier assigns to the earlier author rises by 6.5. What the continuing model inherits is the style of the history, not its text. Restating the earlier turns in a third party's words removes much of the shift, and rewriting the same content in the style of a third model pulls the continuation toward that third model instead. The contagion is directional (from the continuation alone, the author of the history can be picked among ten models at three times the chance rate), it is still present six turns later, and it is not verbatim copying (\S\ref{sec:mech}).

The contagion matters wherever a conversation changes hands, and one such place is the resale market for LLM access. API gateways and resellers (``relays'') sell frontier models at a discount, and what they serve does not always match what they advertise \citep{lin2026gatescope}. Existing audits consider a relay that substitutes a cheaper model for whole requests \citep{cai2025substitution,gao2025met}. We study a relay that forwards the first $\tau$ turns of a conversation to the genuine model, which we call the \emph{claimed model}, and every later turn to a cheaper \emph{substitute}. This is where the money is: the client resends the whole history with every call, so turns 5--8 of an 8-turn dialogue carry 79\% of all input tokens in our data, and a handoff after turn 4 saves 52--61\% of the bill. It is also where the substitute is hardest to see, because it now writes in the claimed model's style.

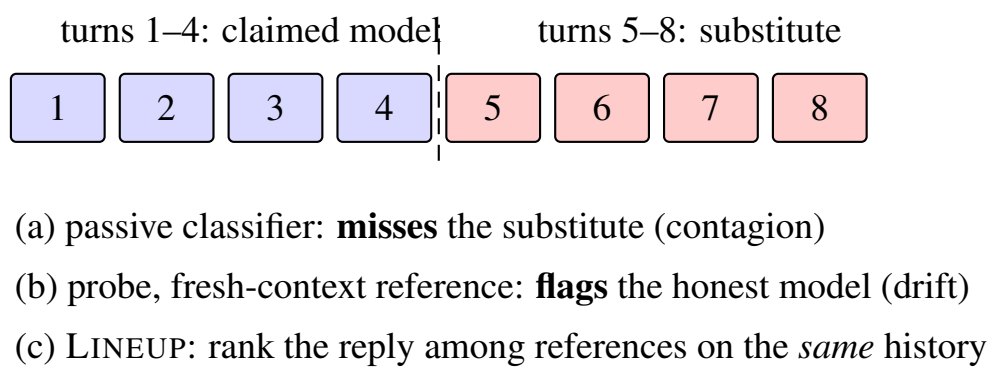
\begin{figure}[t]
\centering
\begin{tikzpicture}[font=\scriptsize, turn/.style={draw, rounded corners=1pt, minimum width=0.62cm, minimum height=0.45cm, inner sep=1pt},
  lab/.style={align=left, anchor=west}]
\foreach \i in {1,...,4} {\node[turn, fill=blue!15] (t\i) at (0.72*\i, 0) {\i};}
\foreach \i in {5,...,8} {\node[turn, fill=red!20] (t\i) at (0.72*\i, 0) {\i};}
\node[anchor=south] at (2.0, 0.3) {turns 1--4: claimed model};
\node[anchor=south] at (4.9, 0.3) {turns 5--8: substitute};
\draw[densely dashed] (3.24,-0.35) -- (3.24,0.55);
\node[lab] at (0.3,-0.8) {(a) passive classifier: \textbf{misses} the substitute (contagion)};
\node[lab] at (0.3,-1.2) {(b) probe, fresh-context reference: \textbf{flags} the honest model (drift)};
\node[lab] at (0.3,-1.6) {(c) \method: rank the reply among references on the \emph{same} history};
\end{tikzpicture}
\caption{Mid-conversation substitution and three audits. The relay forwards turns 1--4 to the claimed model and later turns to a cheaper substitute. Fixed references fail in two directions (a, b) because they ignore the history; \method{} (c) draws the reference from the claimed model on the same history.}
\label{fig:threat}
\end{figure}

Against this relay, audits with a fixed reference fail in two opposite directions (Figure~\ref{fig:threat}):
\begin{itemize}
\item \textbf{Passive detectors miss the substitute.} Text classifiers that identify the serving model from its replies lose much of their power after a handoff (\S\ref{sec:fail1}): detection drops from 0.97 to 0.72 for DeepSeek replacing Claude Sonnet and from 0.87 to 0.54 for a same-vendor downgrade (Haiku). Retraining on handoff data repairs only substitutes seen in training, so it requires knowing the substitute in advance.
\item \textbf{Active probes flag honest relays.} Low-entropy probes (``flip a coin'') are not contagious: they detect every substitute, before and after a handoff. But history changes them in another way. The genuine model's own answers drift on long contexts (\S\ref{sec:fail2}), so an auditor whose reference comes from fresh conversations flags an honest Sonnet in 28\% of long sessions and an honest Gemma in 100\%, and a standard MMD equality test \citep{gao2025met} rejects the honest model for all nine models we test.
\end{itemize}

Both failures have one cause: the reply depends on the history and the reference does not. A fingerprint is a property of a model \emph{and} its history, so the reference should share the history. Our test, \method{} (\S\ref{sec:method}), puts the relay's reply in a lineup: at each audited turn the auditor sends the true history to the official claimed model a few times and ranks the relay's reply among the answers. If the relay is honest, the reply and the references are exchangeable, so the randomized rank is uniform for any score function, per-turn p values combine across turns, and the false-positive rate is controlled by construction, without a fitted threshold, white-box access or log-probabilities. The audited turns are ordinary conversation turns, so the relay cannot recognize the audit. We validate \method{} on white-box proxies (340 honest sessions, unseen substitutes, random per-request replacement, benign serving changes) and, under a preregistered protocol, on two closed models with six substitute pairs: with contemporaneous references the honest false-positive rate is 0.056 and detection after four audited turns is 0.81, for 16 reference calls (about \$0.45 for Sonnet).

\method{} does not undo the contagion, and we do not claim that it does. A passive threshold calibrated on contemporaneous honest sessions of the same configuration, which an auditor rarely has, reaches similar power with a strong score (0.77). What \method{} adds is validity without calibration data, robustness to weak scores, and aggregation over turns. The reference matters more than the statistic: a variant that replaces the rank by an MMD statistic with an embedding kernel detects every substitute session (1.00), at the price of an embedding model.

\paragraph{Contributions.} (1) The finding that model fingerprints are contagious across a conversation, measured with three detectors on ten substitute pairs, and preregistered evidence that the channel is the style of the history. (2) A threat model for mid-conversation substitution with the cost structure that motivates it, and evidence that passive detection fails against it and that retraining does not close the gap for unseen or same-vendor substitutes. (3) Evidence that history also shifts an honest model's own probe answers, so that fresh-context references make probe audits and MMD tests reject honest models, across nine models. (4) The same-history reference principle and \method{}, its simplest instance (any black-box score, one reply per turn, no threshold), with a validity argument and confirmation on local and closed models; an MMD variant is a more powerful instance. All hypotheses of the confirmatory stages were frozen before their data were generated (Appendix~\ref{app:prereg}).

\section{Related Work}

\paragraph{Auditing model substitution.} \citet{cai2025substitution} frame substitution detection for LLM APIs and compare classifiers, identity questions and benchmark-based tests. Model Equality Testing (MET; \citealp{gao2025met}) casts it as a two-sample test with a maximum mean discrepancy (MMD) statistic over a fixed prompt distribution. RUT \citep{zhu2025rut} tests whether API outputs rank uniformly among samples from a local white-box reference model on single-turn prompts. LLMmap \citep{pasquini2025llmmap} fingerprints a model from its answers to a handful of queries, and IRIS \citep{zhang2026iris} audits gateways with budgeted random-generation probes, including routing dilution. GateScope \citep{lin2026gatescope} measures behavioral consistency of API gateways. KBF \citep{fang2026kbf} targets relays with renewable probes near the claimed model's knowledge boundary and calibrates against the reference model's own variation; in our long-history re-implementation it stays valid under its own calibration but drifts under a fresh-context quantile (\S\ref{sec:fail2}); One Token Is Enough \citep{bruckner2026onetoken} fingerprints models by their single-token answer distributions to simple prompts and reports invariance across providers, temperatures and time, measured in fresh contexts. AgentProv \citep{wang2026agentprov} moves to another channel, the distribution of tool calls, with an MMD permutation test, and notes that provider-injected system prompts make text-based tests flag honest providers; LIDAR \citep{wang2026lidar} identifies the model behind a coding agent from its actions. All of these use references that do not depend on a conversation history; we show that this assumption breaks in long conversations, in both directions. \method{} is closest to RUT: it shares the rank statistic, but its reference is the closed claimed model on the audited conversation's own (contaminated) history, its score can be any black-box function, and it aggregates evidence across turns.

\paragraph{Model attribution.} Classifiers recover the source model of a text with high accuracy \citep{sun2025idiosyncrasies}, and recent work targets robustness to unseen domains and model versions \citep{ko2026atlas}. We find that a conversation's history is a strong, previously unexamined confounder: the fingerprint is contagious, and a model continuing another model's conversation is attributed to the other model.

\paragraph{Switching models within an interaction.} Behavior transfers when agents swap models across sessions \citep{choi2026oldhabits}; switching the model of the final turn shifts task performance, with source and target effects that explain about 70\% of the variance \citep{khraishi2026drift}; continuing a non-native trajectory costs agents success \citep{ganz2026handoff}; and interventions on the history causally change later turns \citep{li2026atetheirwords}. These works study benign routing and task outcomes. We study an adversarial setting and the identity signal, and replicate the additive source/target structure for it (Appendix~\ref{app:mech}).

\section{Setting}
\label{sec:setting}

\paragraph{Threat model.} A client sends the full history with every call (the APIs are stateless). The relay answers turns $1..\tau$ with the claimed model and turns $t>\tau$ with a substitute. Metadata such as cache statistics or system fingerprints are returned by the relay and can be forged, so the auditor sees only text. We distinguish two auditors: a \emph{fresh-reference auditor}, whose only reference for the claimed model comes from fresh conversations, and an auditor \emph{with official access}, who can also call the official claimed model. Either auditor can fork a long conversation at any turn and resend its prefix.

\paragraph{Models.} We use 13 models (Table~\ref{tab:models} in Appendix~\ref{app:models}). Claimed models: Claude Sonnet 5.5 and GPT-6.1 Sol (OpenRouter, official providers pinned). Substitutes: the same-vendor downgrades Claude Haiku 5.5 (for Sonnet) and GPT-6 Luna (for Sol); the cross-vendor API models DeepSeek V4.1 Flash (official API) and MiniMax M3; and seven local open-weight models, of which Gemma-3-27B, OLMo-3.1-32B and Qwen3-32B appear in the main experiments and Gemma-3-12B, GLM-4.5-Air, Mistral-Small-24B and Qwen3-8B only in held-out substitute pairs. In the white-box experiments of \S\ref{sec:method}, Gemma-3-27B and OLMo-3.1-32B also play the claimed model.

\paragraph{Conversations.} 200 seed requests from WildChat \citep{zhao2024wildchat} start 8-turn dialogues whose later user turns come from replayed scripts written by a Llama-3.1-8B simulator \citep{grattafiori2024llama3} before any candidate model ran, so user turns never depend on the assistant's replies. Every model has a clean dialogue on every seed; handoff dialogues reuse the claimed model's clean turns $1..\tau$ ($\tau=4$ unless stated) and let the substitute answer the rest. A confirmation stage adds 60 new seeds (code, math, factual QA).

\paragraph{Detectors.} We use three source classifiers. The \emph{n-gram classifier} is a logistic regression on TF-IDF word/character n-grams plus 43 stylometric features. The \emph{neural classifier} is an LLM2Vec encoder \citep{behnamghader2024llm2vec} with a LoRA classification head. The \emph{stylometric classifier} is a logistic regression on the 43 stylometric features only. All three are trained on clean turns to tell apart the same 11 models: five API models (Sonnet, Sol, Haiku, Luna, DeepSeek) and six local models (Gemma-3-12B and -27B, Qwen3-8B and -32B, OLMo-3.1-32B, GLM-4.5-Air). MiniMax M3 and Mistral-Small-24B are not among the classes, so they act as substitutes that the classifiers have never seen. Training uses 5-fold cross-fitting by seed; a seed's text is always scored by the model of its held-out fold. The score of a text for a claimed model is the logit of the probability $P(\text{claimed})$ that the classifier assigns to that model, computed stably as $\log P(\text{claimed}) - \operatorname{logsumexp}_{k\neq \text{claimed}}\log P(k)$. We call the three scores the n-gram, neural and stylometric score.

\section{Failure 1: passive detection misses the substitute}
\label{sec:fail1}

\begin{figure}[t]
\centering
\includegraphics[width=\columnwidth]{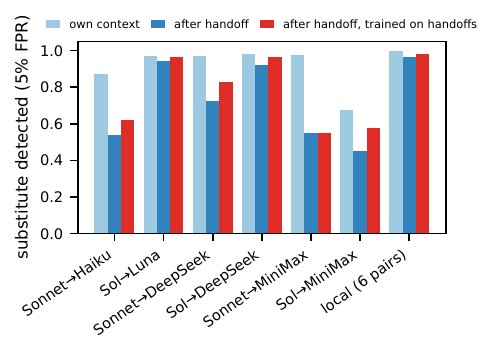}
\caption{Passive detection of the substitute's reply as ``not the claimed model'' (neural classifier, 5\% false-positive rate), in the substitute's own dialogue vs.\ after a handoff from the claimed model, and with a neural classifier retrained on handoff turns. MiniMax is outside the classifier's classes (open world). 140 confirmation seeds; MiniMax 80.}
\label{fig:passive}
\end{figure}

We call the substitute's turn-5 reply ``not the claimed model'' when its score falls below the 5\% quantile of the claimed model's own turn-5 replies (cross-fitted between seed folds; realized false-positive rate 0.06--0.10). Figure~\ref{fig:passive} compares the substitute's replies in its own dialogue with its replies after a handoff from the claimed model.

After the handoff, detection drops from 0.87 to 0.54 for Haiku replacing Sonnet, from 0.97 to 0.72 for DeepSeek replacing Sonnet, and from 0.97 to 0.55 for MiniMax replacing Sonnet (MiniMax is a model the classifiers have never seen); AUROC for Sonnet$\to$Haiku drops from 0.98 to 0.84. The loss depends on the claimed model and on the classifier: with Sol as the claimed model the neural classifier loses little (DeepSeek 0.98$\to$0.92, Luna 0.97$\to$0.94), whereas the weaker n-gram classifier loses heavily for both claimed models (Table~\ref{tab:passive}, Appendix~\ref{app:passive}). Local substitutes remain detectable (1.00$\to$0.96). A natural fix is to train on handoff turns. It repairs substitutes seen in training (Sonnet$\to$DeepSeek 0.72$\to$0.83) but barely helps the same-vendor downgrade (0.54$\to$0.62) and not the unseen substitute (MiniMax 0.55$\to$0.55): retraining requires knowing the substitute in advance.

\section{Mechanism: the contagion is stylistic}
\label{sec:mech}

\begin{figure*}[t]
\centering
\includegraphics[width=\textwidth]{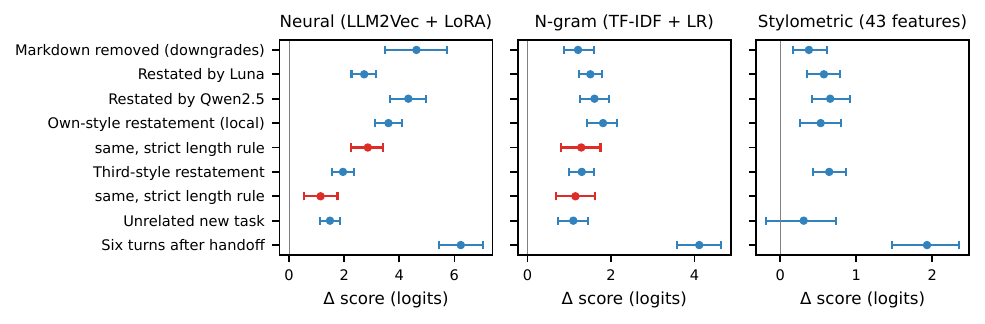}
\caption{Frozen mechanism contrasts on the confirmation sample (95\% seed-cluster bootstrap CIs; semantically flagged paraphrases excluded). Red: stricter exclusion of short paraphrases after the human check. The third-style contrast measures the probability of the third model; all others measure the probability of the claimed model.}
\label{fig:mech}
\end{figure*}

A preregistered mechanism study manipulates the substitute's history while holding the turn-5 request fixed and measures how the score of the substitute's reply changes (Figure~\ref{fig:mech}; 140 confirmation seeds; ten substitute pairs, namely each claimed model with DeepSeek, its same-vendor downgrade and the three main local models; hypotheses frozen after an exploration stage on separate seeds; Appendix~\ref{app:mech}). The handoff itself shifts the substitute's reply toward the claimed model by 6.52 logits [5.93, 7.11] (neural score). Three manipulations isolate the channel:
\begin{itemize}
\item \textbf{Restating the claimed model's turns} by a third-party paraphraser removes much of the shift (2.73 with GPT-6 Luna, 4.33 with Qwen2.5-7B as paraphraser).
\item \textbf{Rewriting the same content in the substitute's own style} reduces the shift of local substitutes (3.61; 2.86 under a stricter exclusion of paraphrases that lose content).
\item \textbf{Rewriting it in a third model's style} moves the substitute toward that third model ($+1.96$ logits of the third model's probability; 1.14 under the stricter rule). Because the history is then written by the third model, not by the substitute, this result is free of the confound that a model may simply prefer its own text.
\end{itemize}
Explicit formatting (markdown) is a second, smaller channel that matters most for same-vendor downgrades: removing it accounts for 63\% of the downgrade shift but 23\% for local substitutes (neural score; 55\% vs.\ 15\% with the n-gram score). The influence is \emph{directional}: from the substitute's replies alone, the source of a history can be guessed among ten models with accuracy 0.30 (chance 0.10). It persists on an unrelated new task (1.48) and six turns after the handoff (6.24), and it is not verbatim copying (for DeepSeek, 1.3\% of its 8-grams occur in the history, and masking them leaves the shift unchanged). All effects have the same sign under the n-gram and stylometric scores; the share of the shift carried by the last history turn was the only preregistered claim not confirmed. A human check of the paraphrases (two annotators, Krippendorff's $\alpha=0.84$ for information loss) finds that 6--18\% lose important content, mostly rewrites in the substitute's own or a third model's style; excluding short paraphrases leaves every conclusion intact and shrinks the own-style and third-style effects by 20--40\%, which we report as the conservative magnitudes.

\section{Failure 2: probes wrongly flag honest relays}
\label{sec:fail2}

Active probes avoid the camouflage. We fork each conversation at turn 5 and ask identity questions, IRIS-style random-number questions \citep{zhang2026iris} (coin, digit, integer, 10-digit string; three repeats each), and LLMmap queries \citep{pasquini2025llmmap}, scoring random-number sessions with a naive-Bayes classifier over the visible answers. At a threshold set on fresh-context sessions, every substitute is detected in 1.00 of sessions whether it answers in its own context, after the handoff, or three turns later (0.89--1.00 with thresholds calibrated on the honest model's answers to the same kind of history); the claimed model's history does not camouflage it (preregistered contrast: difference $-0.006$ [$-0.025$, 0.013]). Identity self-reports are useless in either direction: DeepSeek calls itself ChatGPT in 64\% of fresh conversations, and a relay-style system prompt makes every substitute claim to be the claimed model.

\begin{figure}[t]
\centering
\includegraphics[width=\columnwidth]{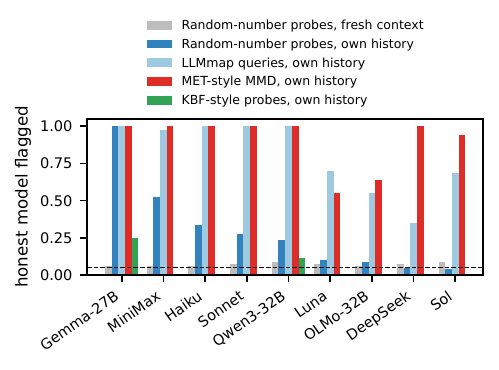}
\caption{A fresh-reference auditor's false positives on \emph{honest} models answering on their own long history (threshold: 5\% quantile of fresh-context sessions, cross-fitted). Random-number probes are scored by the parsed value only. MET: MMD two-sample test with a character Hamming kernel, fresh-context reference vs.\ own-history samples, 100 repetitions (fresh vs.\ fresh: 0.01--0.11). KBF: knowledge-boundary probes, Gemma and Qwen3 only (80 sessions each).}
\label{fig:drift}
\end{figure}

The problem is on the honest side. Treating each of nine models in turn as the claimed model (the six API models, Gemma-27B, OLMo-32B and Qwen3-32B), Figure~\ref{fig:drift} measures how often its own answers on its own long history are flagged by a threshold set on its fresh-context answers. With random-number probes, five of nine models exceed 0.15 (Gemma-27B 1.00, MiniMax 0.53, Haiku 0.34, Sonnet 0.28, Qwen3-32B 0.24), and the drift is in the random values themselves---scoring only the parsed number gives the same rates. Gemma's ``random digit'' moves from 3 (94\%) to 7 (77\%); Haiku's ``random integer'' collapses onto 37. LLMmap queries are flagged at 0.35--1.00, partly because answers start to refer to the conversation. A standard MET test \citep{gao2025met} with fresh-context references rejects the honest model in 55--100\% of repetitions for every model. Knowledge-boundary probes (KBF, \citealp{fang2026kbf}; our reduced re-implementation, Appendix~\ref{app:kbf}) drift less: with the same fresh-context threshold an honest Gemma is flagged in 25\% and Qwen3 in 11\% of long sessions (preregistered contrast: $+0.12$ [0.06, 0.18] over fresh sessions), and KBF's own rule, a binomial test against an upper confidence bound of the reference's self-discrepancy, keeps the honest rate at 0.03 while flagging 0.97 of substitutes after a handoff. Fresh-context references thus turn a long conversation into evidence of substitution. An auditor with official access can avoid this by calibrating on official answers \emph{to similar histories}, but most probes have a second weakness: they look like probes. A TF-IDF classifier trained on probe families recognizes 98\% of identity, random-number, knowledge and alignment probes at 1\% false positives on real user requests, so a relay can route recognized probes to the claimed model (Appendix~\ref{app:probeid}). KBF-style factual cloze probes are the exception (0.00 when unseen, 0.40 after one example), which makes KBF the strongest probe audit for a fresh-reference auditor; it needs a per-model enrollment, which failed for OLMo (6 stable facts).

\section{\texorpdfstring{\method}{Lineup}: testing against a same-history reference}
\label{sec:method}

\paragraph{Test.} At an audited turn $t$ with true history $H_t$ (including any replies already produced by the relay) and user message $x_t$, the auditor sends $(H_t,x_t)$ to the official claimed model $m$ times and receives references $y^{(1)},\dots,y^{(m)}$. With a score function $s$ and the relay's reply $y$,
\begin{align*}
u_t &= \frac{1}{m+1}\Big(\#\{j: s(y^{(j)}) < s(y)\}\\
&\qquad + V\,\big(\#\{j: s(y^{(j)})=s(y)\}+1\big)\Big),
\end{align*}
with $V\sim U(0,1)$.
Small $u_t$ means the reply is less like the claimed model than the claimed model's own answers to the same history. A session of $k$ audited turns is rejected when Fisher's statistic $-2\sum_t \log u_t$ \citep{fisher1925} exceeds the $1-\alpha$ quantile of $\chi^2_{2k}$. We call this test \method{}: the relay's reply stands in a lineup with $m$ replies of the claimed model to the same history, and the test asks whether it stands out.

The test assumes that the relay's reply and the references are generated with the same sampling parameters and the same reasoning effort; the auditor fixes this configuration for its reference calls.
\begin{proposition}[validity]
\label{prop:validity}
If the relay forwards every audited turn to the claimed model with the configuration used for the references, then $y,y^{(1)},\dots,y^{(m)}$ are exchangeable given $(H_t,x_t)$, so $u_t\sim U(0,1)$ for any score function $s$. Since fresh references are drawn at every audited turn given the true history, $u_t$ is uniform given all earlier $u$'s; the Fisher statistic is $\chi^2_{2k}$ and the session-level false-positive rate equals $\alpha$.
\end{proposition}
\begin{proposition}[single-turn power bound]
\label{prop:bound}
For any alternative, $P(u_t\le\alpha)\le\min\{1,(m+1)\alpha\}$; power therefore requires aggregation across turns.
\end{proposition}
$u_t$ is a conformal p value \citep{vovk2005alrw}; the propositions follow from exchangeability and the uniformity of the randomized rank. The null includes the serving configuration, and the references must be sampled close in time to the audited replies (\S\ref{sec:closed}). Because the audited turns are ordinary conversation turns and the references go to the official API, the relay can neither recognize the audit nor see the references. As black-box scores we use the neural and n-gram scores of \S\ref{sec:setting}; on white-box proxies only, we also use the mean token log-likelihood under the claimed model (the likelihood score).

\subsection{Local proxies}

White-box models stand in for the claimed model (Gemma-27B, OLMo-32B) with substitutes in handoff sessions (turns 5--8 audited, 80 test seeds per pair; Appendix~\ref{app:local}).

\paragraph{Validity.} Over 340 honest sessions the per-turn $u$ is uniform for all three scores (Kolmogorov--Smirnov $p$ 0.15--0.48; $P(u<0.05)$ 0.047--0.054) and the session false-positive rate is 0.02--0.07 at $\alpha=0.05$ and 0.00--0.02 at $\alpha=0.01$.

\begin{figure}[t]
\centering
\includegraphics[width=\columnwidth]{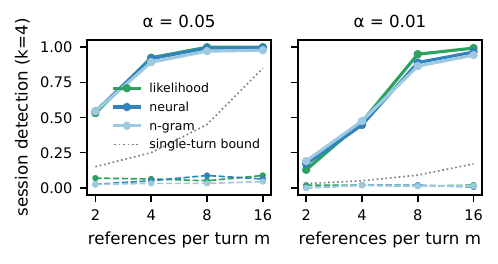}
\caption{Local proxies: detection by \method{} after $k=4$ audited turns vs.\ references per turn (solid, mean of four substitute pairs) and false-positive rate of honest sessions (dashed).}
\label{fig:costpower}
\end{figure}

\paragraph{Cost and power.} Figure~\ref{fig:costpower} shows detection after four audited turns. With $m=4$ references per turn detection is 0.89--0.95 at $\alpha=0.05$ for the neural and likelihood scores (0.76--0.95 for the n-gram score); $\alpha=0.01$ needs $m=8$ (0.81--0.97). Single-turn detection stays near the bound of Proposition~\ref{prop:bound}. In 16-turn conversations detection reaches 1.00 by $k=8$ for handoffs after turn 4 or 8 (likelihood and neural scores), while honest sessions (pooled over claimed models and scores) stay at 0.00--0.07 up to $k=12$ (Appendix~\ref{app:e7}).

\paragraph{Unseen substitutes (preregistered).} When the substitute is not among the scorer's classes, \method{} beats a fixed threshold calibrated on honest sessions of the other fold: $+0.09$ [0.05, 0.13] in session-paired detection over three new pairs and their applicable scorers ($p<0.001$). The gain concentrates where the scorer is weak or unfamiliar with the substitute: Gemma$\to$Mistral with the n-gram score, 0.71 vs.\ 0.48.

\paragraph{Diluted substitution and benign changes.} If the relay replaces each later turn independently with probability $\varepsilon$ and the auditor combines all audited turns, detection after eight turns is 0.35--0.36 for $\varepsilon=0.25$, 0.78--0.79 for $\varepsilon=0.5$ and 1.00 for full substitution, while honest sessions stay at 0.01--0.11 (Figure~\ref{fig:rr}, Appendix). Benign serving changes---a different sampling temperature, top-$p$ truncation, an added system prompt, FP8 weights---are rejected at rates close to a pure-randomness control under the neural score (0.00--0.14 vs.\ 0.05--0.07); the white-box likelihood score, in contrast, rejects a temperature change in every session (Appendix~\ref{app:local}). The score function therefore decides what counts as a deviation: classifier scores test which model serves, likelihoods test the exact serving configuration.

\subsection{Closed models}
\label{sec:closed}

We confirm \method{} on Sonnet and Sol with six substitute pairs (Sonnet with Haiku, DeepSeek or MiniMax; Sol with Luna, DeepSeek or MiniMax), under hypotheses frozen before generation: 80 honest sessions per claimed model, 40 sessions per pair, turns 5--8 generated now by the claimed model or the substitute on the claimed model's pilot history, and $m=4$ references from the official model after each turn, so replies and references are contemporaneous. Reasoning cannot be disabled for either closed model; we use low reasoning effort for every reply and reference of a model.

\begin{figure}[t]
\centering
\includegraphics[width=\columnwidth]{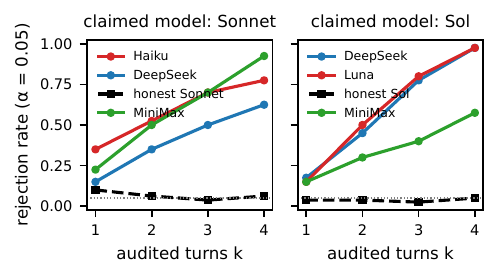}
\caption{Closed claimed models: rejection rate of \method{} (neural score, $m=4$, $\alpha=0.05$) vs.\ audited turns; honest sessions dashed (n=80 per model, 40 per substitute pair).}
\label{fig:closed}
\end{figure}

\begin{table}[t]
\centering
\small
\setlength{\tabcolsep}{4pt}
\begin{tabular}{llcccc}
\toprule
 & & \multicolumn{2}{c}{passive threshold} & \multicolumn{2}{c}{\method} \\
Score & Sessions & $k=1$ & $k=4$ & $k=1$ & $k=4$ \\
\midrule
Neural & substitutes & 0.55 & 0.77 & 0.25 & \textbf{0.82} \\
 & honest & 0.07 & 0.09 & 0.07 & 0.05 \\
N-gram & substitutes & 0.38 & 0.55 & 0.20 & \textbf{0.75} \\
 & honest & 0.06 & 0.07 & 0.06 & 0.06 \\
\bottomrule
\end{tabular}
\caption{Closed models, passive detection vs.\ \method{} on the same sessions ($m=4$, $\alpha=0.05$; post hoc analysis, not preregistered). Rows show the detection rate for substitute sessions and the false-positive rate for honest sessions. The passive threshold is the 5\% quantile of contemporaneous honest turns ($k=1$) or summed scores ($k=4$) of the other fold, its most favorable setting. Per-pair results in Table~\ref{tab:closed}.}
\label{tab:e1}
\end{table}

The honest false-positive rate is 0.056 [0.025, 0.094] (preregistered: one-sided upper bound 0.088 $<$ 0.10, Holm $p=0.025$), with uniform per-turn ranks (Kolmogorov--Smirnov $p=0.77$). Detection over all substitute sessions is 0.81 [0.76, 0.85] (preregistered: lower bound 0.77 $>$ 0.5). Figure~\ref{fig:closed} breaks this down by turn and pair.

\paragraph{What \method{} adds.} Table~\ref{tab:e1} compares it with passive detection on the same sessions. The camouflage costs both alike: against a threshold calibrated on contemporaneous honest sessions of the same configuration---a favorable baseline that an auditor rarely has---\method{} is on par under the strong neural score (0.82 vs.\ 0.77; on the same-vendor downgrade 0.78 vs.\ 0.70, a preregistered advantage that was not confirmed, $p=0.23$). With a single audited turn the passive threshold is stronger (0.55 vs.\ 0.25): by Proposition~\ref{prop:bound}, one turn with $m=4$ references cannot reject more than $(m+1)\alpha=0.25$ of sessions. More references per turn raise the four-turn detection of \method{}: at $k=4$ it is 0.29, 0.51 and 0.82 for $m=1$, 2 and 4. The gains of \method{} are elsewhere: it needs no calibration data, it keeps its power with the weaker n-gram score while the threshold does not (0.75 vs.\ 0.55; Sol$\to$MiniMax 0.53 vs.\ 0.07), and it aggregates turns. On the local proxies, where the substitute's own-context replies can be audited too, the handoff does not lower \method's detection (neural score: 0.88 own context vs.\ 0.92 after the handoff; Appendix~\ref{app:revision}).

\paragraph{Score and statistic.} The principle is the same-history reference, and \method{} is its simplest instance. \method-MMD, which replaces the rank by an MMD statistic with an RBF kernel on the neural classifier's embeddings and within-turn permutations, rejects every closed substitute session at $k=4$ (0.86 at $k=2$) with honest false-positive rate 0.05; with MET's character kernel it reaches only 0.58, and it cannot reject at $k=1$ (five permutations per turn). The score of \method{} matters as well: with Stouffer's combination, averaging the standardized neural and n-gram scores gives 0.92 (neural alone 0.89, stylometric 0.67). By the Neyman--Pearson lemma the most powerful rank statistic against a known substitute is the likelihood ratio $\log p_{\text{claimed}}(y\mid H_t,x_t)-\log p_{\text{substitute}}(y\mid H_t,x_t)$; the neural score, $\log P(y\mid\text{claimed})-\log\sum_{k\neq\text{claimed}}P(y\mid k)$ under uniform priors, is a learned approximation against a mixture of known models, which is why it tests \emph{which} model serves, whereas $\log p_{\text{claimed}}$ alone tests any deviation from the claimed model's configuration (\S\ref{sec:method}). Stouffer's combination is best under full substitution, Fisher's under diluted substitution ($\varepsilon=0.5$: 0.79 vs.\ 0.68--0.78), and the minimum p value (Tippett) is weak throughout; the combination should be fixed in advance (Appendix~\ref{app:revision}).

\begin{figure}[t]
\centering
\includegraphics[width=\columnwidth]{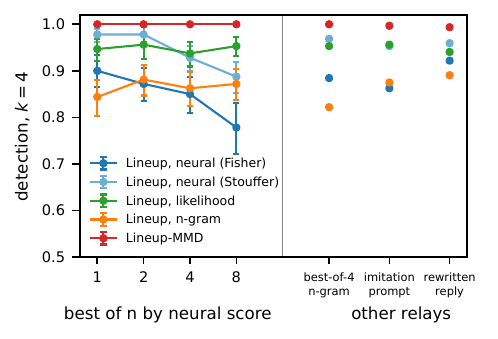}
\caption{Adaptive relays on the local proxies: detection after $k=4$ turns ($m=4$, $\alpha=0.05$, 320 sessions per condition, 95\% seed-cluster CIs). Conditions: best of $n$ substitute candidates by the neural score; best of 4 by the n-gram score; imitation system prompt; reply rewritten by Qwen3-8B toward the claimed model's earlier replies.}
\label{fig:attack}
\end{figure}

\paragraph{Score-aware relays (preregistered).} A relay that knows the audit score can sample $n$ candidates from the substitute and return the one that looks most like the claimed model. On the local proxies (four pairs, 80 seeds each, hypotheses frozen after an exploration on separate seeds), best-of-8 selection by the neural score lowers the detection of \method{} with that score from 0.90 to 0.78 ($-0.12$ [$-0.18$, $-0.06$]; Figure~\ref{fig:attack}), most for OLMo$\to$Mistral (0.91$\to$0.57). Selecting by a different score (n-gram, best of 4) lowers detection under the neural score less than selecting by the neural score itself, but not significantly (0.88 vs.\ 0.85; not confirmed, $p=0.08$). The attack does not transfer to the likelihood score, which keeps 0.95 (the drop under the neural score exceeds the drop under the likelihood score by 0.13 [0.06, 0.19]), nor to \method-MMD, which detects every session under every attack; Stouffer's combination retains 0.89 and $m=8$ references 0.93. Imitation prompts and a rewrite of the substitute's reply by a small model toward the claimed model's earlier replies cost less than best-of-8. The attack multiplies the substitute's output cost by $n$. Keeping the score private, or auditing with a statistic that uses more than one direction of the reply (likelihood, embeddings), defeats it.

\paragraph{Timing and cost.} Timing matters: in an earlier run whose references were sampled two to four days after the audited replies, 3 of 24 honest Sonnet sessions were rejected at $k=4$; contemporaneous sampling restores the nominal rate. An audit of four turns costs 16 reference calls, about \$0.45 for Sonnet and \$0.12 for Sol.

\section{Discussion}

\paragraph{Which audit to use.} Table~\ref{tab:methods} summarizes the trade-offs. Passive classifiers are free but miss camouflaged substitutes; probes are robust to camouflage but recognizable and, with fresh references, unfair to honest relays; calibrating probes on official answers to similar histories fixes the second problem but not the first. \method{} needs $m\times k$ official calls per audited session but no threshold, no white-box access and no special requests. Auditors with official access should prefer it for long conversations. Three practical rules follow from our results. (i) Sample the references concurrently with the audited turn (same configuration, same minute), and when an audit spans a provider update, rerun it rather than reuse old references; under rate limits, audit fewer sessions rather than widen the gap between reply and references. (ii) Choose the score for what should count as a deviation: classifier scores (neural, or n-gram plus neural) for ``which model serves'', likelihoods only when the exact configuration is part of the contract; keep the score private. (iii) Fix the combination in advance (Stouffer for full, Fisher for diluted substitution). Per-turn conformal p values can also be turned into test martingales for continuous, anytime-valid auditing \citep{vovk2005alrw,shafer2021betting,ramdas2023gametheoretic}; with $m=4$ references, however, a mixture power martingale detects far fewer substitutes than Fisher at a fixed horizon (closed models, $k=4$: 0.09 vs.\ 0.82; Appendix~\ref{app:e7}), so we recommend fixing the audit length in advance.

\begin{table*}[t]
\centering
\small
\setlength{\tabcolsep}{3pt}
\newcolumntype{L}[1]{>{\raggedright\arraybackslash}p{#1}}
\begin{tabular}{@{}L{3.8cm}L{4.0cm}L{3.5cm}L{2.5cm}L{1.3cm}@{}}
\toprule
Method & Camouflaged substitute & Honest model on long history & Requires & Visible to relay \\
\midrule
Passive classifier, fixed threshold & missed (0.54--0.72, $k=1$) & ok & text only & no \\
\quad + handoff training & seen substitutes only & ok & handoff data of the substitute & no \\
\quad contemporaneous calibration & 0.77 (closed, $k=4$) & ok (0.09) & honest sessions of same config. & no \\
Random-number probes, fresh reference & detected (1.00) & flagged (0.24--1.00 in 5/9 models) & fresh samples & yes (0.98) \\
Random-number probes, official matched reference & detected (0.89--1.00) & ok & official calls & yes \\
MET-style MMD, fresh reference & -- & flagged (0.55--1.00, 9/9 models) & fresh samples & yes \\
KBF-style probes, own calibration & detected (0.97) & ok (0.03); fresh quantile 0.11--0.25 & per-model enrollment & hard (0.00--0.40) \\
\method{} (same-history rank test) & detected (0.81 closed, 0.9--1.0 local; $k=4$) & ok (0.056) & $m\times k$ official calls & no \\
\quad vs.\ best-of-8 relay that knows the neural score & 0.78 (neural), 0.95 (likelihood); local & -- & -- & -- \\
\method-MMD, embedding kernel & detected (1.00 closed, $k=4$; 1.00 vs.\ best-of-8) & ok (0.05) & + embedding model, $k\ge 2$ & no \\
\bottomrule
\end{tabular}
\caption{Audit methods for mid-conversation substitution, with the evidence from this paper.}
\label{tab:methods}
\end{table*}

\section{Limitations}

We study chat; coding agents, the most common use of relays, have long system prompts and execute tools. A pilot with a 2k-token agent-style system prompt and tool definitions keeps validity (honest 0.03--0.06) and detection (0.81--0.93 at $k=4$) on local proxies, but its tools were never executed and only 1\% of replies called one (Appendix~\ref{app:e7}). We audit two closed claimed models, and the substitutes are a convenience set of current models. The closed-model confirmation uses low reasoning effort for both replies and references; the default configuration was tested only in a smaller earlier run (Appendix~\ref{app:e7}). The null of \method{} includes the serving configuration, so an honest relay with a different sampling setup may be rejected by likelihood-based scores (not by classifier scores in our tests), and references must be sampled close in time to the audited replies because providers are not stationary; we did not measure how long references stay exchangeable with the replies (a two- to four-day gap was too long in one run), so contemporaneous sampling is a premise of the test, not a measured window. A relay that knows the score can lower the power of \method{} by best-of-$n$ selection (0.90$\to$0.78 at $n=8$ on local proxies), so the score should stay private; we did not test such relays on closed models or attackers that optimize the likelihood or embedding statistics. Single-turn power is bounded, so the test needs several audited turns per session, and it costs $m\times k$ official calls. Our users are replayed simulator scripts; an interactive simulator gave the same handoff effects (Appendix~\ref{app:mech}), and with real WildChat user turns the handoff shift keeps its sign but is smaller for two of four local pairs, while \method{} keeps its validity and power (Appendix~\ref{app:e7}). Our style-transfer experiment with third-party editors failed to produce a measurable style difference and is reported as a negative result; the style claim rests on the contrast in which the history is rewritten in a third model's style.

\section*{Ethics Statement}
Our findings could help relays hide substitution (e.g., hand off later, keep the claimed model's text in the history). We report them together with an audit that does not depend on the substitute being recognizable, and we do not evaluate or name commercial relays. The human check of paraphrases was carried out by two annotators: the author and another researcher, who took part voluntarily and without compensation. They judged 100 model-generated paraphrases for lost, added and changed content, and saw only model-generated text and seed requests from a public dataset.

\bibliography{references}

\appendix

\section{Preregistration and deviations}
\label{app:prereg}
Each confirmatory stage froze its hypotheses, statistics, directions, thresholds, samples and exclusion rules in a configuration file committed before its data were generated. The hypothesis IDs below (H1a--H23) are those of the released configuration files; they are used only in this appendix and in the table of Appendix~\ref{app:mech}, and the main text describes each hypothesis in words. No hypotheses carry the IDs H12 and H13. The four original stages are the mechanism study (H1a--H7, Holm across nine), the probe stage (H8--H11), the local-proxy stage (H14) and the closed-model stage (H15--H17). Results of the frozen tests: mechanism 8/9 under the neural score (the API part of H4, the recency share, not confirmed); probes: H8 (a handoff camouflages probes) not confirmed, H9 confirmed but negligible (+1.2 points), H10 and H11 not confirmed (laundering is unnecessary for probes and shifts the honest model); H14 (\method{} beats a fixed threshold on unseen substitutes) confirmed; H15 (false-positive bound) and H16 (detection) confirmed, H17 (advantage on the same-vendor downgrade) not confirmed. In a later revision round, two further stages were frozen before their data: adaptive relays (H18--H20, Holm across three; H18, best-of-8 selection lowers detection, and H20, the attack does not transfer to the likelihood score, confirmed; H19, selection by a different score, not confirmed) and KBF-style probes (H21--H23; all confirmed). In the KBF stage, OLMo could not serve as a claimed model (6 stable facts), which the frozen configuration records. Deviations are listed in the released reports, e.g.\ the number of references $m=4$ fixed by budget, and a temperature change of 0.7$\to$1.0 for the local proxies whose official temperature is 0.7.

\section{Mechanism study details}
\label{app:mech}
\paragraph{Conditions.} All conditions keep the turn-5 user message and change only the substitute's history (turns 1--4). \emph{Original}: the claimed model's original replies. \emph{No markdown}: the same with markdown removed. \emph{Third-party restatement}: restated by GPT-6 Luna or by Qwen2.5-7B with the paraphrase prompt of \citet{sun2025idiosyncrasies}. \emph{Own-style restatement}: restated by the substitute itself. \emph{Third-style restatement}: restated by a third model. \emph{Own-style with original last turn}: the own-style restatement with the claimed model's original last turn. \emph{New task}: the claimed model's history, or the substitute's own history, followed by an unrelated new task. Control: the substitute's own clean dialogue at the same turn. Paraphrases flagged by an automatic semantic check (embedding similarity, length ratio, truncation, refusals) are excluded per condition, as frozen.

\paragraph{Confirmation results.} 140 seeds (80 held out from the exploration stage, 60 new tasks), ten substitute pairs, one-sided seed-cluster bootstrap, Holm across nine hypotheses.

\begin{table*}[t]
\centering
\small
\begin{tabular}{llll}
\toprule
ID & Contrast & Neural & N-gram \\
\midrule
H1a & Markdown removed, downgrades & 4.62 [3.49, 5.73] & 1.21 [0.87, 1.59] \\
H1b & \quad downgrades $-$ local & 3.56 [2.28, 4.81] & 0.50 [$-$0.03, 1.05]$^\dagger$ \\
H2 & Third-party restatement & 2.73 / 4.33 & 1.51 / 1.60 \\
H3a & Own-style restatement, local & 3.61 [3.11, 4.09] & 1.81 [1.43, 2.15] \\
H3b & Third-style restatement & 1.96 [1.57, 2.34] & 1.30 [1.00, 1.59] \\
H4 & Recency share, local / API & 0.32 / 0.41$^\dagger$ & 0.21 / 0.37$^\dagger$ \\
H5 & Unrelated new task & 1.48 [1.11, 1.85] & 1.10 [0.73, 1.45] \\
H6 & 10-way source accuracy & 0.30 (chance 0.10) & 0.26 \\
H7 & Six turns after handoff & 6.24 [5.43, 7.03] & 4.11 [3.57, 4.63] \\
\bottomrule
\end{tabular}
\caption{Frozen mechanism hypotheses on the confirmation sample (change in score, in logits; $^\dagger$ not confirmed). The first two rows compare the original history with the no-markdown history; the third-style row measures the shift toward the third model. Under the stricter paraphrase rule the own-style contrast is 2.86 / 1.29 and the third-style contrast 1.14 / 1.15 (neural / n-gram).}
\end{table*}

\paragraph{Held-out pairs.} The handoff shift reproduces on all four held-out pairs, including a new family (Sonnet$\to$Mistral 9.72, Sol$\to$Mistral 4.16, Sonnet$\to$GLM 5.07, Sol$\to$Gemma-12B 5.27, neural score). Transfer to an unrelated task (new task after the claimed model's history minus new task after the substitute's own history) is weak on held-out pairs and on most local substitutes; it is carried by the API and downgrade substitutes.

\paragraph{Directional structure.} Across a 10$\times$9 source--substitute matrix, an additive model with a source ``contagiousness'' and a substitute ``susceptibility'' term explains 62\% of the variance under the neural score and 32\% under the n-gram score, consistent with the $\sim$70\% reported for task performance by \citet{khraishi2026drift}. Closed models are hard to sway: in reverse handoffs Sonnet moves toward a local source by 3.98 logits and Sol by 0.21, against 6.8--12.9 for local substitutes.

\paragraph{Dynamics.} The shift decays within one to two turns after the handoff and stays above zero up to turn 12; replacing the claimed model's original turns by the substitute's own restatement after the handoff removes most of the persistent part for local substitutes. Within a reply, the beginning is the most like the claimed model. With an interactive user simulator in place of replayed user turns the effect has the same size, and the simulated user turns themselves do not reveal whether a handoff happened (AUROC 0.49--0.52).

\paragraph{Human check of paraphrases.} Two annotators judged 100 paraphrases (50 flagged, 50 unflagged by the automatic check) for lost, added and changed content. Agreement was good for lost content (Krippendorff's $\alpha=0.84$), moderate for added content (0.68) and low for changed meaning (0.60), where the annotators disagreed on hedging changes. Weighted to the population, 6\% (both annotators) to 18\% (either) of the paraphrases lose important content; the automatic check catches 17--38\% of the severe cases, and short paraphrases (length ratio $<0.9$) catch most of the rest. The stricter exclusion leaves all conclusions intact.

\paragraph{A failed style-transfer design.} We also tried a content$\times$style 2$\times$2 design in which third-party editors (GLM-4.5-Air, OLMo-32B) rewrote each history turn in the claimed model's or the substitute's style from three reference passages. The edits did not move style: the two cells with the same content differed by $\le$1.5 logits while the originals differed by 15--24, and edits drifted toward the editor's own style (38\% of OLMo's edits were attributed to OLMo). The design failed its preregistered manipulation gate and was not run; few-shot rewriting erases the source style without importing the target style, in line with the third-party restatement result.

\section{Passive detection: all detectors}
\label{app:passive}
\begin{table}[h]
\centering
\small
\setlength{\tabcolsep}{3pt}
\begin{tabular}{@{}lcccccc@{}}
\toprule
 & \multicolumn{3}{c}{N-gram} & \multicolumn{3}{c}{Neural} \\
Handoff & own & after & +train & own & after & +train \\
\midrule
Sonnet$\to$Haiku & 0.51 & 0.24 & 0.32 & 0.87 & 0.54 & 0.62 \\
Sonnet$\to$DeepSeek & 0.87 & 0.45 & 0.78 & 0.97 & 0.72 & 0.83 \\
Sol$\to$DeepSeek & 0.83 & 0.43 & 0.72 & 0.98 & 0.92 & 0.96 \\
Sol$\to$Luna & 0.51 & 0.38 & 0.52 & 0.97 & 0.94 & 0.96 \\
Sonnet$\to$MiniMax$^*$ & 0.94 & 0.60 & 0.57 & 0.97 & 0.55 & 0.55 \\
Sol$\to$MiniMax$^*$ & 0.41 & 0.19 & 0.24 & 0.68 & 0.45 & 0.57 \\
\bottomrule
\end{tabular}
\caption{Detection of the substitute's turn-5 reply as ``not the claimed model'' (5\% nominal false-positive rate, realized 0.06--0.12). \emph{own}: in the substitute's own dialogue; \emph{after}: after the handoff; \emph{+train}: after the handoff, with handoff turns added to the classifier's training data. $^*$Outside the classifier's classes.}
\label{tab:passive}
\end{table}
Treating DeepSeek as unseen by refitting the n-gram classifier without its class gives 0.45$\to$0.21 detection after the handoff, the same drop as for MiniMax.

\section{Probe identifiability}
\label{app:probeid}
Positives: 407 probes in four families---identity (90), random generation (180), model knowledge (72), alignment and opinion (65)---from our probes, LLMmap's queries and IRIS-style prompts, with paraphrases. Negatives: 5{,}000 unused English first turns from WildChat and 1{,}600 replayed user turns. A TF-IDF (word and character n-grams) logistic regression recognizes 0.98 of probes at 1\% false positives when it has seen all families; embedding the probe in a natural request lowers this to 0.46. Unseen families are harder (identity 0.60, knowledge 0.46, alignment 0.30, random generation 0.04), but one example of a family suffices to recognize it. KBF-style cloze probes (80 batches of ten items, 40 single items) are the exception: 0.00 when unseen and 0.40 after one example (0.18 when embedded in a natural request). None of the 320 audited user turns of our local-proxy sessions was recognized as a probe, so a probe-aware relay does not affect \method.

\section{Local proxies: additional results}
\label{app:local}
\begin{figure}[h]
\centering
\includegraphics[width=\columnwidth]{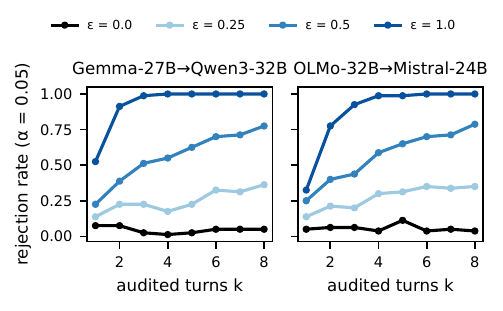}
\caption{Random per-request replacement with probability $\varepsilon$ (\method{} with the neural score, $m=8$).}
\label{fig:rr}
\end{figure}
\begin{figure}[h]
\centering
\includegraphics[width=\columnwidth]{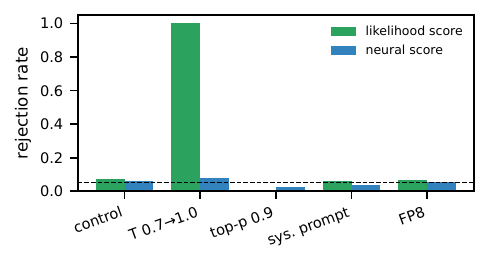}
\caption{Rejection of benign serving changes by \method{} ($k=4$, $\alpha=0.05$; mean of two proxies).}
\label{fig:benign}
\end{figure}

\section{KBF-style probes}
\label{app:kbf}
We re-implemented the core of KBF \citep{fang2026kbf} for local models. Enrollment: the reference model proposes ``name $|$ value'' facts in six numeric domains (boiling and melting points, chromosome numbers, years, summit elevations, drug half-lives; 3 rounds of 15 per domain); a fact is kept if the model's six answers under three prompt variants at $T=0.7$ agree within a domain tolerance (118 facts for Gemma and Qwen3, 6 for OLMo). Audit: forked at turn 5, five batches of ten cloze items; a session's discrepancy count is tested against a binomial with $p_0$, the one-sided 99\% Clopper--Pearson bound of the reference's discrepancy rate in 40 fresh calibration sessions (0.26 Gemma, 0.23 Qwen3). Confirmation (80 seeds): honest own-history sessions flagged 0.00 (Gemma) and 0.05 (Qwen3); fresh sessions 0.00; substitutes after the handoff 0.89--1.00 (mean discrepancy 0.38--0.61 vs.\ 0.19--0.27 for the honest model). With the Figure~\ref{fig:drift} threshold (95\% quantile of fresh sessions of the other fold) honest own-history sessions are flagged 0.25 and 0.11.

\section{Extensions (descriptive, not preregistered)}
\label{app:e7}
Designs fixed before generation (\texttt{configs/revision\_e7.json}); local proxies only (the four local pairs of \S\ref{sec:method}, $m=4$, $\alpha=0.05$, Fisher).

\begin{table}[h]
\centering
\small
\setlength{\tabcolsep}{3.5pt}
\resizebox{\columnwidth}{!}{%
\begin{tabular}{llcccc}
\toprule
Setting & Sessions & $k=1$ & $k=4$ & $k=8$ & $k=12$ \\
\midrule
16 turns, honest & 80 & .06 & .01 & .00 & .03 \\
16 turns, handoff $\tau=4$ & 160 & .20 & .87 & 1.00 & 1.00 \\
16 turns, handoff $\tau=8$ & 160 & .22 & .89 & .99 & -- \\
Agent prompt, honest & 80 & .05 & .03 & -- & -- \\
Agent prompt, handoff & 160 & .19 & .81 & -- & -- \\
Real users, honest & 120 & .06 & .07 & -- & -- \\
Real users, handoff & 240 & .25 & .85 & -- & -- \\
\bottomrule
\end{tabular}}
\caption{\method{} with the neural score in long conversations, with an agent-style system prompt and with real user turns. Likelihood score: 1.00 at $k=8$ (16 turns), 0.93 (agent) and 0.96 (real users) at $k=4$, honest 0.03--0.06.}
\label{tab:e7}
\end{table}

\paragraph{Long conversations.} The replayed scripts were extended to 16 turns with the pilot's simulator method (40 seeds). For the honest sessions, per-turn ranks are uniform in each block of four turns for all three scores (Kolmogorov--Smirnov $p\ge0.27$, $P(u<0.05)$ 0.047--0.059) except the neural score in turns 9--12 ($p=0.005$, one of nine tests; $P(u<0.05)=0.05$, no excess of small ranks). The n-gram score is the weakest for OLMo$\to$Mistral (0.88--0.93 at $k=8$--12).

\paragraph{Agent-style system prompt.} A 2k-token system prompt (assistant persona, policies, five tool definitions with a JSON call format) precedes every request of both the claimed model and the substitute and every reference (40 seeds). Tools are never executed; 1\% of replies contain a tool call.

\paragraph{Real users.} 60 English WildChat conversations with at least eight user turns (non-toxic, not redacted) are replayed verbatim; these turns originally responded to ChatGPT replies. The handoff shift of the substitute's reply toward the claimed model (neural score, turns 5--8, handoff minus the substitute's own dialogue) keeps its sign but differs in size from the simulator scripts: 0.99 vs.\ 1.02 (Gemma$\to$Qwen3), 1.99 vs.\ 2.71 (Gemma$\to$OLMo), 4.37 vs.\ 8.41 (OLMo$\to$Gemma), 2.35 vs.\ 8.06 (OLMo$\to$Mistral).

\paragraph{Anytime-valid auditing.} With the mixture power martingale $M_k=\int_0^1\prod_{t\le k}\kappa\, u_t^{\kappa-1}\,d\kappa$ and rejection at $M_k\ge 1/\alpha$, honest sessions cross in 0.00--0.03 of sessions, but power is far below Fisher at the same horizon: closed models ($m=4$, $k=4$) 0.09 vs.\ 0.82; random replacement ($m=8$, $k=8$) 0.96 vs.\ 1.00 for $\varepsilon=1$ and 0.34 vs.\ 0.78 for $\varepsilon=0.5$; 16-turn handoffs at $\tau=4$ 0.93 vs.\ 1.00 (neural score, $k=12$), mean stopping turn of rejected sessions 3.0--7.2. With few references the per-turn p values are too coarse for the martingale to grow quickly.

\paragraph{Reasoning effort.} The earlier closed-model run used OpenRouter's default reasoning for Sonnet (955 reasoning tokens per reference on average, vs.\ 254 at low effort in the confirmation) and drew its references two to four days after the replies. At $k=4$ it detected 0.92 of Haiku and DeepSeek substitutes (24 sessions each) and rejected 3 of 24 honest sessions; the two changes are confounded, and we attribute the excess to the time gap (\S\ref{sec:closed}).

\section{Post hoc analyses (not preregistered)}
\label{app:revision}
The analyses in this appendix reuse data of the preregistered stages. They were added in a later revision round, after the confirmatory results were known; their scripts and comparisons were fixed before running, but they are not preregistered. The one exception is Table~\ref{tab:closed}, which breaks the frozen closed-model analysis down by pair. Randomized ranks are redrawn per analysis, which moves detection rates by about $\pm0.03$ (e.g.\ 0.81 in the frozen detection test, 0.82 in Table~\ref{tab:e1}).

\begin{table}[h]
\centering
\small
\begin{tabular}{lrr}
\toprule
 & \method & passive, $k=4$ \\
\midrule
Sonnet, honest & 0.06 & 0.06 \\
Sonnet$\to$Haiku & 0.78 & 0.70 \\
Sonnet$\to$DeepSeek & 0.62 & 0.62 \\
Sonnet$\to$MiniMax & 0.93 & 0.95 \\
Sol, honest & 0.05 & 0.11 \\
Sol$\to$Luna & 0.97 & 0.93 \\
Sol$\to$DeepSeek & 0.97 & 0.97 \\
Sol$\to$MiniMax & 0.57 & 0.42 \\
\bottomrule
\end{tabular}
\caption{Closed models per pair, $k=4$, $\alpha=0.05$, neural score (frozen closed-model analysis). Honest rows show false-positive rates. The passive threshold is calibrated on contemporaneous honest sessions of the same configuration (other fold).}
\label{tab:closed}
\end{table}

\paragraph{Camouflage cost on local proxies.} For the four local pairs we also audited the substitute in its own clean dialogue, with $m=4$ references from the claimed model on that same history, and compared with the handoff sessions (first four of eight references). Detection at $k=4$ (Fisher) is 0.93 vs.\ 0.95 (likelihood score), 0.91 vs.\ 0.90 (n-gram score) and 0.88 vs.\ 0.92 (neural score) for own context vs.\ handoff; the only drop is OLMo$\to$Mistral under the n-gram score (0.88$\to$0.76). Local substitutes are camouflaged little in the first place (passive detection 1.00$\to$0.96); the closed substitutes, which are camouflaged most, cannot be audited in their own context without new calls.

\begin{table*}[t]
\centering
\small
\begin{tabular}{lcccc}
\toprule
 & \multicolumn{2}{c}{closed} & \multicolumn{2}{c}{local} \\
Test & $k=2$ & $k=4$ & $k=2$ & $k=4$ \\
\midrule
\method-MMD, RBF on neural embeddings & .03 / .86 & .05 / 1.00 & .05 / .91 & .05 / 1.00 \\
\method-MMD, character Hamming & .03 / .30 & .08 / .58 & .04 / .33 & .06 / .67 \\
\method{}, neural score & .04 / .42 & .05 / .82 & -- & .05 / .92 \\
\bottomrule
\end{tabular}
\caption{Same-history tests, honest false-positive rate / substitute detection, $m=4$ references per turn, $\alpha=0.05$. MMD null: within-turn permutation of the reply among its references (1{,}000 draws); RBF bandwidth: median pairwise distance.}
\label{tab:mmd}
\end{table*}

\begin{table}[h]
\centering
\small
\begin{tabular}{lccc}
\toprule
 & \multicolumn{2}{c}{closed} & local \\
Score & honest & substitute & substitute \\
\midrule
N-gram & .07 / .07 & .76 / .81 & .89 / .97 \\
Neural & .04 / .06 & .79 / .89 & .92 / .98 \\
Stylometric & .06 / .05 & .61 / .67 & .78 / .84 \\
N-gram + neural & .06 / .05 & .85 / .92 & .93 / .99 \\
Likelihood & -- & -- & .94 / 1.00 \\
Likelihood + neural & -- & -- & .97 / 1.00 \\
\bottomrule
\end{tabular}
\caption{Score ablation for \method{}, $k=4$, $m=4$, Fisher / Stouffer. Combined scores average scores standardized with constants frozen on the training folds.}
\label{tab:scores}
\end{table}

\paragraph{Combination.} On random per-request replacement (neural score, $m=8$) with $\varepsilon=1$, $k=4$, Fisher, Stouffer and Cram\'er--von Mises reach 0.99 and Tippett 0.36; with $\varepsilon=0.5$, $k=8$, Fisher reaches 0.79 for both pairs, Stouffer 0.68--0.78, Cram\'er--von Mises 0.56--0.61 and Tippett 0.17--0.23; honest sessions stay at 0.01--0.06.

\section{Models, prices and detectors}
\label{app:models}
\begin{table}[h]
\centering
\small
\resizebox{\columnwidth}{!}{%
\begin{tabular}{lllll}
\toprule
Model & Vendor & Weights & \$/M in / out & Class \\
\midrule
Claude Sonnet 5.5$^\dagger$ & Anthropic & closed & 2.00 / 10.00 & yes \\
GPT-6.1 Sol$^\dagger$ & OpenAI & closed & 2.00 / 10.00 & yes \\
Claude Haiku 5.5 & Anthropic & closed & 0.10 / 0.50 & yes \\
GPT-6 Luna & OpenAI & closed & 0.10 / 0.50 & yes \\
DeepSeek V4.1 Flash & DeepSeek & API & 0.15 / 0.60 & yes \\
MiniMax M3 & MiniMax & API & 0.30 / 1.20 & no \\
Gemma-3-12B/27B & Google & open & -- & yes \\
OLMo-3.1-32B & AI2 & open & -- & yes \\
Qwen3-8B/32B & Alibaba & open & -- & yes \\
GLM-4.5-Air & Zhipu & open (FP8) & -- & yes \\
Mistral-Small-24B & Mistral & open & -- & no \\
\bottomrule
\end{tabular}}
\caption{Models ($^\dagger$claimed models) and list prices at the time of the experiments (OpenRouter, official providers; DeepSeek through its official API). Savings of a handoff are computed from these prices and the measured token counts. \emph{Class}: whether the model is one of the 11 classes of the source classifiers. Qwen2.5-7B (paraphraser) and Llama-3.1-8B (user simulator) produce no audited replies and are not listed.}
\label{tab:models}
\end{table}
\paragraph{Detectors.} N-gram classifier: word 1--2-gram (30k features) and character 2--4-gram (50k) TF-IDF with sublinear term frequency, plus 43 standardized stylometric features (length, punctuation, markdown, lexical statistics), multinomial logistic regression with $C$ chosen by inner cross-validation (100 in every fold). Neural classifier: LLM2Vec encoder (Llama-3.1-8B-Instruct, MNTP and supervised adapters; \citealp{behnamghader2024llm2vec}) with bidirectional attention, a trainable LoRA and a linear head over mean-pooled states, first 512 tokens, 3 epochs, batch 8, learning rate $5\times10^{-5}$ with cosine schedule and 10\% warmup, weight decay 0.001, gradient clipping 0.3. Stylometric classifier: logistic regression ($C=1$ by inner CV) on the 43 stylometric features. All are 11-way classifiers over the models marked in Table~\ref{tab:models}, cross-fitted over five seed folds.
\paragraph{Paraphrase filters.} A paraphrase is flagged when its cosine similarity to the original (BGE-large embeddings) is below 0.85, its length ratio is outside 0.5--1.5, it is truncated, or it is a refusal; a sample is additionally checked with an NLI model (RoBERTa-large-MNLI) in both directions. Leading and trailing remarks about the paraphrase (``Here is a paraphrased version:'') are stripped before use.
\paragraph{Code and data.} Code, the frozen configuration files, reports, generation logs (provider, model snapshot, token counts) and analysis scripts are available at \url{https://github.com/TTtianTT/TurnPrint}.

\end{document}